\documentclass[10pt,twocolumn,letterpaper]{article}

\usepackage{cvpr}
\usepackage{times}
\usepackage{epsfig}
\usepackage{graphicx}
\usepackage{amsmath}
\usepackage{amssymb}
\usepackage[percent]{overpic}
\usepackage{times}
\usepackage{epsfig}
\usepackage{graphicx}
\usepackage{float}
\usepackage{wrapfig}
\usepackage{amsmath,amssymb,amsthm}
\usepackage{algorithm,algorithmicx,algpseudocode}
\usepackage{bm,xspace}
\usepackage{comment}
\usepackage{verbatim}
\usepackage{multirow}
\usepackage{balance}
\usepackage{url}
\usepackage{booktabs}
\usepackage{etoolbox,siunitx}
\usepackage{calc}
\usepackage{pifont,hologo}

\usepackage[pagebackref=true,breaklinks=true,letterpaper=true,colorlinks,bookmarks=false]{hyperref}

\cvprfinalcopy 

\def\cvprPaperID{1011} 

\ifcvprfinal\fi
\begin{document}

\title{Guided Stereo Matching}

\author{Matteo Poggi\thanks{Joint first authorship.} , Davide Pallotti$^*$, Fabio Tosi, Stefano Mattoccia\\
Department of Computer Science and Engineering (DISI)\\
University of Bologna, Italy\\
{\tt\small \{m.poggi, fabio.tosi5, stefano.mattoccia \}@unibo.it}
}

\makeatletter
\g@addto@macro\@maketitle{
  \begin{figure}[H]
  \setlength{\linewidth}{\textwidth}
  \setlength{\hsize}{\textwidth}
  \vspace{-7mm}
  \centering
  \renewcommand{\tabcolsep}{1pt} 
 	\begin{tabular}{ccc}
        \includegraphics[width=0.33\textwidth]{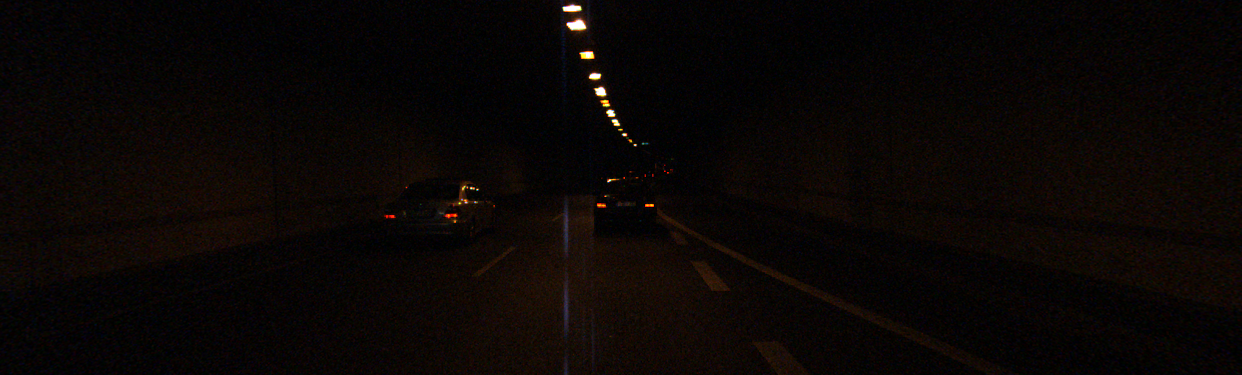} 
        &
        \begin{overpic}[width=0.33\textwidth]{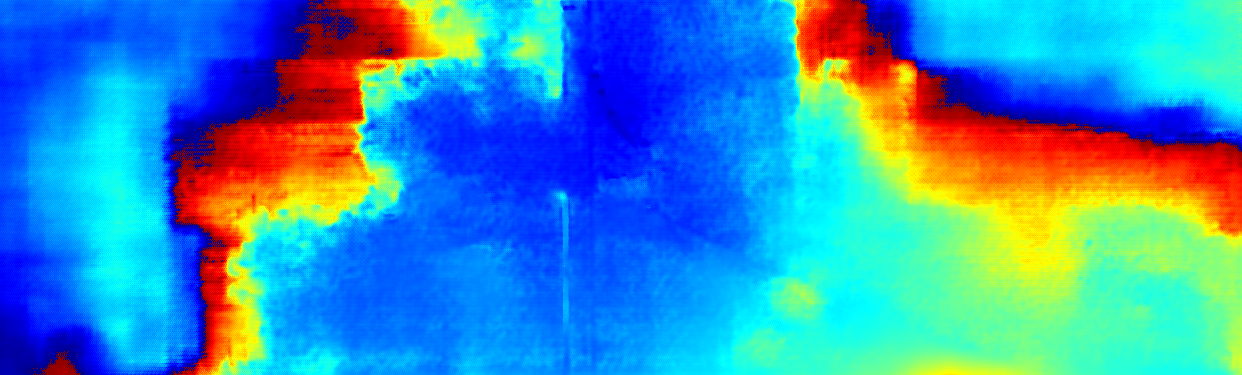}
        \put (4,25) {$\displaystyle\textcolor{white}{\textbf{93.21\%}}$}
        \end{overpic} 
        &
        \begin{overpic}[width=0.33\textwidth]{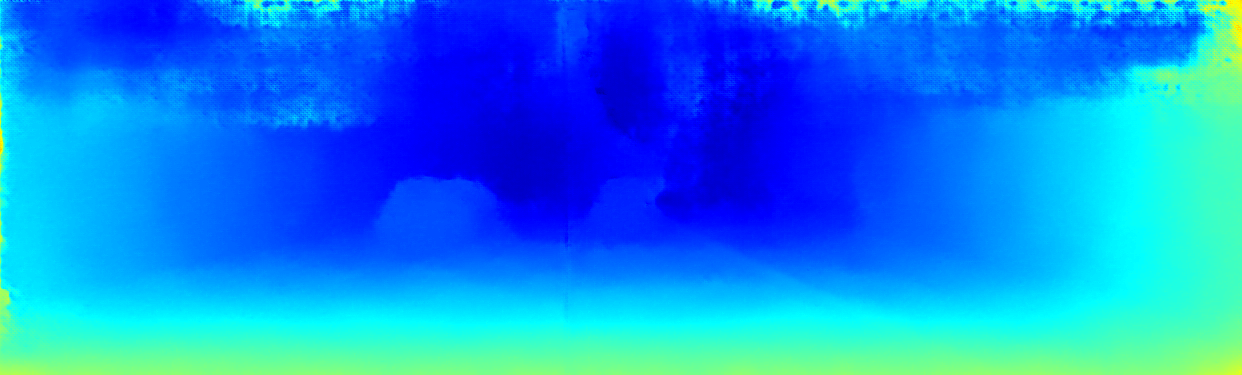}
        \put (4,25) {$\displaystyle\textcolor{white}{\textbf{4.36\%}}$}
        \end{overpic} \\
        (a) & (b) & (c) \\
    \end{tabular}
    \vspace{1mm}
    \caption{\textbf{Guided stereo matching.} (a) Challenging, reference image from KITTI 2015 \cite{KITTI_2015} and disparity maps estimated by (b) iResNet \cite{Liang_2018_CVPR} trained on synthetic data \cite{Mayer_2016_CVPR}, or (c) guided by sparse depth measurements (5\% density). Error rate ($>3$) superimposed on each map.}
  \label{fig:abstract}
  \end{figure}
}
\makeatother

\maketitle

\begin{abstract}

Stereo is a prominent technique to infer dense depth maps from images, and deep learning further pushed forward the state-of-the-art, making end-to-end architectures unrivaled when enough data is available for training. However, deep networks suffer from significant drops in accuracy when dealing with new environments. Therefore, in this paper, we introduce Guided Stereo Matching, a novel paradigm leveraging a small amount of sparse, yet reliable depth measurements retrieved from an external source enabling to ameliorate this weakness. The additional sparse cues required by our method can be obtained with any strategy (\eg, a LiDAR) and used to enhance features linked to corresponding disparity hypotheses.
Our formulation is general and fully differentiable, thus enabling to exploit the additional sparse inputs in pre-trained deep stereo networks as well as for training a new instance from scratch. Extensive experiments on three standard datasets and two state-of-the-art deep architectures show that even with a small set of sparse input cues, i) the proposed paradigm enables significant improvements to pre-trained networks. Moreover, ii) training from scratch notably increases accuracy and robustness to domain shifts. Finally, iii) it is suited and effective even with traditional stereo algorithms such as SGM.

\end{abstract}

\section{Introduction}

Obtaining dense and accurate depth estimation is pivotal to effectively address higher level tasks in computer vision such as autonomous driving, 3D reconstruction, and robotics. It can be carried out either employing active sensors, such as LiDAR, or from images acquired by standard cameras. The former class of devices suffers from some limitations, depending on the technology deployed to infer depth. For instance, sensors based on structured light have limited working range and are ineffective in outdoor environments, while LiDARs, although very popular and accurate, provide only sparse depth measurements and may have shortcomings when dealing with reflective surfaces. On the other hand, passive sensors based on standard cameras potentially allow obtaining dense depth estimation in any environment and application scenario.
Stereo \cite{scharstein2002taxonomy} relies on two (or more) rectified images to compute the disparity by matching corresponding pixels along the horizontal epipolar line, thus enabling to infer depth via triangulation. The most recent trend in stereo consists in training end-to-end Convolutional Neural Networks (CNNs) on a large amount of (synthetic) stereo pairs \cite{Mayer_2016_CVPR} to directly infer a dense disparity map.  However, deep stereo architectures suffer when \emph{shifting domain}, for example moving from synthetic data used for the initial training \cite{Mayer_2016_CVPR} to the real target imagery.  Therefore, deep networks are fine-tuned in the target environment to ameliorate domain shift issues. Nonetheless, standard benchmarks used to assess the accuracy of stereo \cite{KITTI_2012,KITTI_2015,MIDDLEBURY_2014,ETH3D} give some interesting insights concerning such paradigm. While it is unrivaled when a massive amount of data is available for fine-tuning, as is the case of KITTI datasets \cite{KITTI_2012,KITTI_2015}, approaches mixing learning and traditional pipelines \cite{taniai2018continuous} are still competitive with deep networks when not enough data is available, as particularly evident with Middlebury v3 \cite{MIDDLEBURY_2014} and ETH3D \cite{ETH3D} datasets. 

In this paper, we propose to leverage a small set of sparse depth measurements to obtain, with deep stereo networks, dense and accurate estimations in any environment. It is worth pointing out that our proposal is different from depth fusion strategies (\eg, \cite{Marin_ECCV_2016,nair2013survey,dal2015probabilistic,agresti2017deep}) aimed at combining the output of active sensors and stereo algorithms such as Semi-Global Matching \cite{hirschmuller2005accurate}. Indeed, such methods mostly aim at selecting the most reliable depth measurements from the multiple available using appropriate frameworks whereas our proposal has an entirely different goal. In particular, given a deep network and a small set (\eg, less than 5\% of the whole image points) of accurate depth measurements obtained by any means: can we improve the overall accuracy of the network without retraining? Can we reduce domain shift issues? Do we get better results training the network from scratch to exploit sparse measurements?
To address these goals, we propose a novel technique acting at feature level and deployable with any state-of-the-art deep stereo network. Our strategy enhances the features corresponding to disparity hypotheses provided by the sparse inputs maintaining the stereo reasoning capability of the original deep network. It is versatile, being suited to boost the accuracy of pre-trained models as well as to train a new instance from scratch to achieve even better results. Moreover, it can also be applied to improve the accuracy of conventional stereo algorithms like SGM. In all cases, our technique adds a negligible computational overhead to the original method. 
It is worth noting that active sensors, especially LiDAR-based, and standard cameras are both available as standard equipment in most autonomous driving setups. Moreover, since the cost of LiDARs is dropping and solid-state devices are already available \cite{SOLID_STATE_LIDAR}, sparse and accurate depth measurement seems not to be restricted to a specific application domain. Thus, independently of the technology deployed to infer sparse depth data, to the best of our knowledge this paper proposes the first successful attempt to leverage an external depth source to boost the accuracy of state-of-the-art deep stereo networks. We report extensive experiments conducted with two top-performing architectures with source code available, PSMNet by Chang et al.\ \cite{Chang_2018_CVPR} and iResNet by Liang et al.\ \cite{Liang_2018_CVPR}, and standard datasets KITTI\cite{KITTI_2012,KITTI_2015}, Middlebury v3 \cite{MIDDLEBURY_2014} and ETH3D \cite{ETH3D}. The outcome of such evaluation supports the three following main claims of this work:

\begin{itemize}
    \item Given sparse ($<$ 5\% density) depth inputs, applying our method to pre-trained models always boosts its accuracy, either when the network is trained on synthetic data only or fine-tuned on the target environments.
    
    \item Training from scratch a network guided by sparse inputs dramatically increases its generalization capacity, significantly reducing the gap due to domain shifts (\eg, when moving from synthetic to real imagery).
    
    \item The proposed strategy can be applied seamlessly even to conventional stereo algorithms such as SGM.
    
\end{itemize}

In Figure \ref{fig:abstract} we can notice how on a very challenging stereo pair from KITTI 2015 \cite{KITTI_2015} (a) a state-of-the-art deep stereo network trained on synthetic data produces inaccurate disparity maps (b), while guiding it with our method deploying only 5\% of sparse depth data allows for much more accurate results (c) despite the domain shift.

\begin{figure*}
    \centering
    \renewcommand{\tabcolsep}{1pt}   
    \begin{tabular}{ccccc}

            \includegraphics[width=0.19\textwidth]{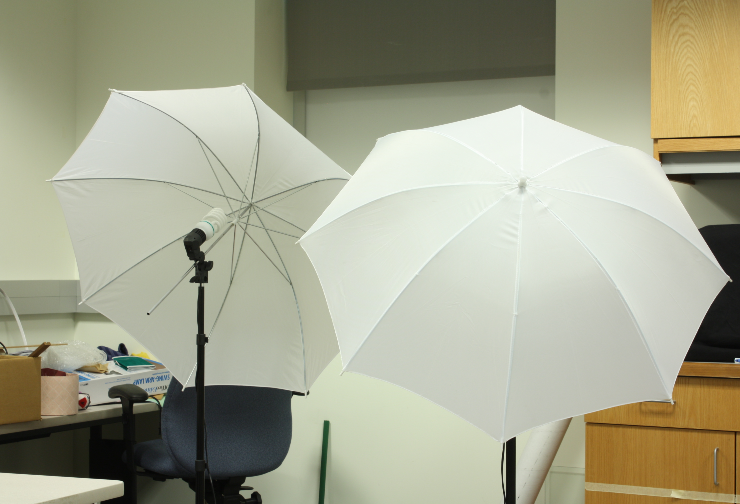} 
            &
            \begin{overpic}[width=0.19\textwidth]{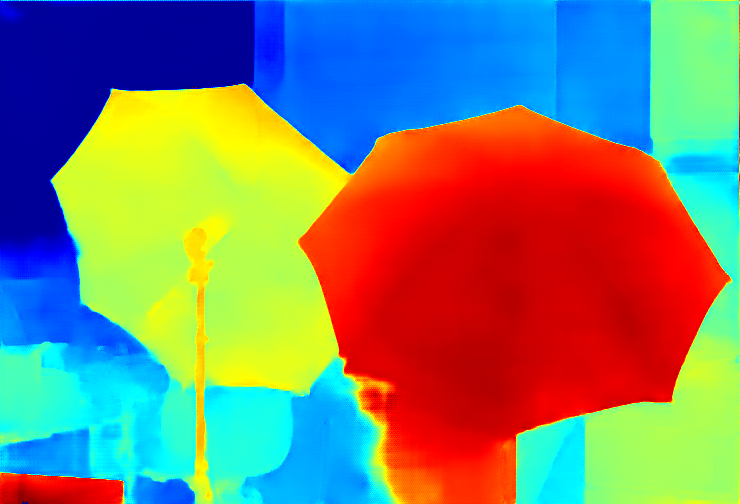}
            \put (4,58) {$\displaystyle\textcolor{white}{\textbf{avg = 3.364}}$}
            \end{overpic} 
            &
            \begin{overpic}[width=0.19\textwidth]{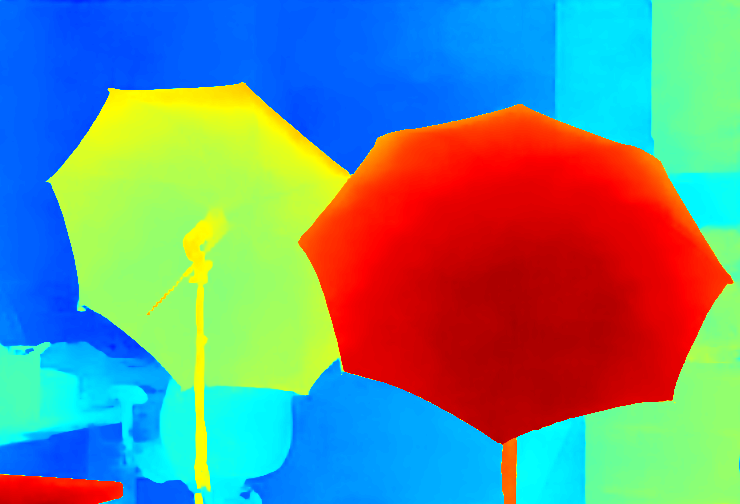}
            \put (4,58) {$\displaystyle\textcolor{white}{\textbf{avg = 0.594}}$}
            \end{overpic} 
            &
            \begin{overpic}[width=0.19\textwidth]{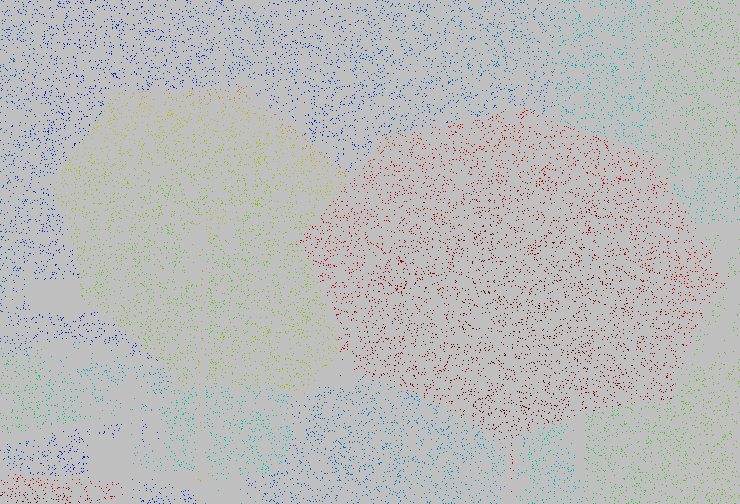}
            \put (4,58) {$\displaystyle\textcolor{black}{\textbf{density = 3.37\%}}$}
            \end{overpic} 
            &
            \includegraphics[width=0.19\textwidth]{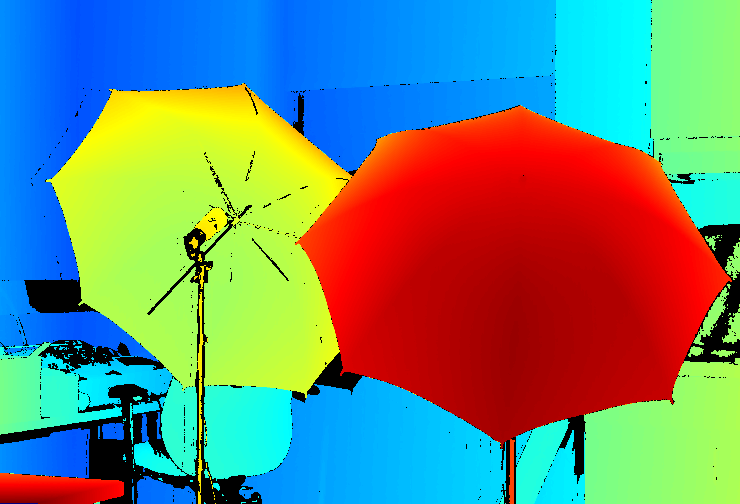}
            \\
            (a) & (b) & (c) & (d) & (e)\\
    \end{tabular}
    \caption{\textbf{Example of improved generalization.} (a) Reference image from Middlebury \cite{MIDDLEBURY_2014}, disparity maps obtained by (b) iResNet \cite{Liang_2018_CVPR} trained on SceneFlow synthetic dataset \cite{Mayer_2016_CVPR}, (c) iResNet trained on SceneFlow for guided stereo, (d) visually enhanced sparse depth measurements taken from (e) ground-truth depth. We stress the fact that (b) and (c) are obtained training on synthetic images only.}
    \label{fig:umbrella}
\end{figure*}

\section{Related work}

Stereo has a long history in computer vision and Scharstein and Szeliski \cite{scharstein2002taxonomy} classified conventional algorithms into two main broad categories, namely \emph{local} and \emph{global} approaches, according to the different steps carried out: i) cost computation, ii) cost aggregation, iii) disparity optimization/computation and iv) disparity refinement. 
While local algorithms are typically fast, they are ineffective in the presence of low-texture regions. On the other hand, global algorithms perform better at the cost of higher complexity. Hirschmuller's SGM \cite{hirschmuller2005accurate} is often the favorite trade-off between the two worlds and for this reason the preferred choice for most practical applications.
Early attempts to leverage machine learning for stereo aimed at exploiting learning-based confidence measures \cite{Poggi_2017_ICCV} to detect outliers or improve disparity accuracy \cite{Spyropoulos_2014_CVPR,Park_2015_CVPR,Poggi_2016_3DV}. Some works belonging to the latter class modified the \emph{cost volume}, an intermediate representation of the matching relationships between pixels in the two images, by replacing winning matching costs \cite{Spyropoulos_2014_CVPR} or modulating their distribution \cite{Park_2015_CVPR} guided by confidence estimation.

The spread of deep learning hit stereo matching as well. Early works focused on a single step of traditional stereo pipelines, for example learning a matching function by means of CNNs \cite{zbontar2015computing,Chen_2015_ICCV,luo2016efficient}, improving optimization carried out by SGM \cite{Seki_2016_BMVC,Seki_2017_CVPR} or refining disparity maps \cite{Gidaris_2017_CVPR,batsos2018recresnet}. 
Later, the availability of synthetic data \cite{Mayer_2016_CVPR} enabled to train end-to-end architectures for disparity estimation embodying all the steps mentioned above. In the last year, a large number of frameworks appeared, reaching higher and higher accuracy on KITTI benchmarks \cite{KITTI_2012,KITTI_2015}. All of them can be broadly categorized into two main classes according to how they represent matching relations between pixels along the epipolar line, similarly to what cost volume computation does for traditional stereo algorithms.

The first class consists of networks computing correlation scores between features belonging to the left and right frames. The outcome are feature maps, linked to disparity hypotheses, concatenated along the channel dimension. This volume is processed through 2D convolutions, usually by encoder-decoder architectures. DispNetC by Mayer et al.\ \cite{Mayer_2016_CVPR} was the first end-to-end network proposed in the literature suggesting this paradigm. More recent architectures such as CLR \cite{Pang_2017_ICCV_Workshops}, iResNet \cite{Liang_2018_CVPR}, DispNet3 \cite{ilg2018occlusions} were built on top of DispNetC. In addition, other frameworks such as EdgeStereo \cite{song2018stereo} and SegStereo \cite{yang2018segstereo} jointly tackled stereo with other tasks, respectively edge detection and semantic segmentation. 

The second class consists of frameworks building 3D cost volumes (actually, 4D considering the feature dimension) obtained by concatenation \cite{Kendall_2017_ICCV} or difference \cite{khamis2018stereonet} between left and right features. Such data structure is processed through 3D convolutions, and the final disparity map is the result of a differentiable \emph{winner-takes-all} (WTA) strategy. GC-Net by Kendall et al.\ \cite{Kendall_2017_ICCV} was the first work to follow this strategy and the first end-to-end architecture reaching the top of the KITTI leaderboard. Following architectures built upon GC-Net improved accuracy, adding specific aggregation modules \cite{yu2018kandao} and spatial pyramidal pooling \cite{Chang_2018_CVPR}, or efficiency, by designing a tiny model \cite{khamis2018stereonet}.

Despite the different strategies adopted, both classes somehow encode the representation of corresponding points in a data structure similar to the cost volume of conventional hand-crafted stereo algorithms. Therefore, with deep stereo networks and conventional algorithm, we will act on such data structure to guide disparity estimation with sparse, yet accurate depth measurements. 

\section{Guided stereo matching}

Given sparse, yet precise depth information collected from an external source, such as a LiDAR or any other means, our principal goal is to exploit such cues to assist state-of-the-art deep learning frameworks for stereo matching. 
For this purpose, we introduce a \emph{feature enhancement} technique, acting directly on the intermediate features processed inside a CNN by peaking those directly related to the depth value suggested by the external measurement. This can be done by precisely acting where an equivalent representation of the cost volume is available. 
The primary goal of such an approach is to further increase the reliability of the already highly accurate disparity map produced by CNNs. Moreover, we also aim at reducing the issues introduced by domain shifts. By feeding sparse depth measurements into a deep network during training, we also expect that it can learn to exploit such information together with image content, compensating for domain shift if such measurements are fed to the network when moving to a completely different domain (\eg, from synthetic to real imagery). Our experiments will highlight how, following this strategy, given an extremely sparse distribution of values, we drastically improve the generalization capacity of a CNN. Figure \ref{fig:umbrella} shows how deploying a 3.36\% density of sparse depth inputs is enough to reduce the average error of iResNet from 3.364 to 0.594.

\subsection{Feature enhancement}

Traditional stereo algorithms collect into a cost volume the relationship between potentially corresponding pixels across the two images in a stereo pair, either encoding similarity or dissimilarity functions. The idea we propose consists in opportunely acting on such representation, still encoded within modern CNNs employing correlation or concatenation between features from the two images, favoring those disparities suggested by the sparse inputs. Networks following the first strategy \cite{Mayer_2016_CVPR,Liang_2018_CVPR,yang2018segstereo,song2018stereo,Pang_2017_ICCV_Workshops} use a \emph{correlation layer} to compute similarity scores that are higher for pixels that are most likely to match, while networks based on the second strategy rely upon a 3D volume of concatenated features.
The cost volume of a conventional stereo algorithm has dimension $H \times W \times D$, with $H \times W$ being the resolution of the input stereo pair and $D$ the maximum disparity displacement considered, while representative state-of-the-art deep stereo networks rely on data structures of dimension, respectively, $H \times W \times (2D+1)$ \cite{Mayer_2016_CVPR} and $H \times W \times D \times 2F$ \cite{Kendall_2017_ICCV}, being $F$ the number of features from a single image.
Given sparse depth measurements $z$, we can easily convert them into disparities $d$ by knowing the focal length $f$ and baseline $b$ of the setup used to acquire stereo pairs, as $d = b \cdot f \cdot \frac{1}{z}$.


With the availability of sparse external data in the disparity domain, we can exploit them to peak the correlation scores or the features activation related to the hypotheses suggested by these sparse hints and to dampen the remaining ones. For example, given a disparity value of $k$, we will enhance the $k$\emph{-th} channel output of a correlation layer or the $k$\emph{-th} slice of a 4D volume.
For our purposes, we introduce two new inputs, both of size $H \times W$: a (sparse) matrix $G$, conveying the externally provided disparity values, and a binary mask $V$, specifying which elements of $G$ are valid (\ie, if $v_{ij}=1$). For each pixel with coordinates $(i,j)$ in the reference image such that $v_{ij}=1$ we alter features as discussed before, based on the known disparity value $g_{ij}$. On the other hand, every point with $v_{ij}=0$ is left untouched. Thus, we rely on the ability of the deep network to reason about stereo and jointly leverage the additional information conveyed by sparse inputs.

\begin{figure*}
    \begin{tabular}{cc}
        \includegraphics[width=0.48\textwidth]{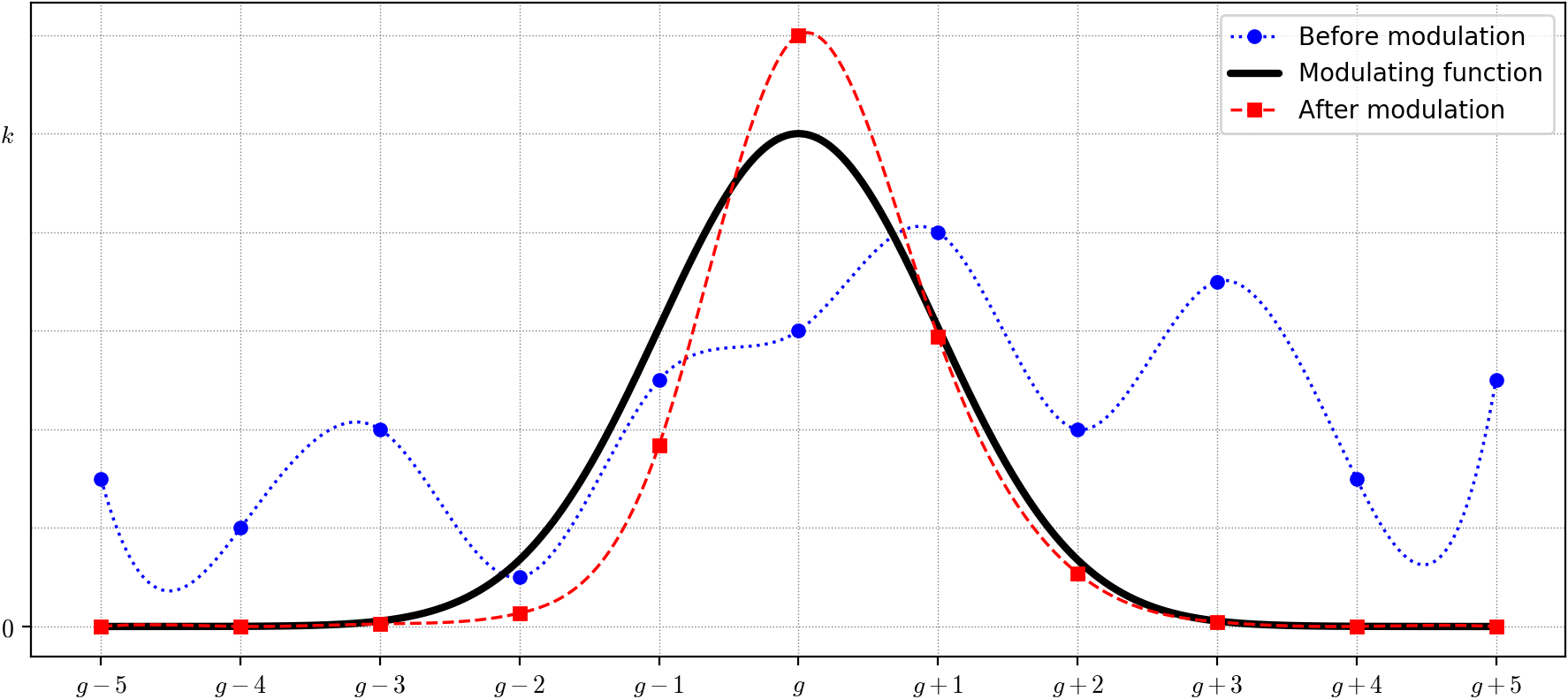}
        &
        \includegraphics[width=0.48\textwidth]{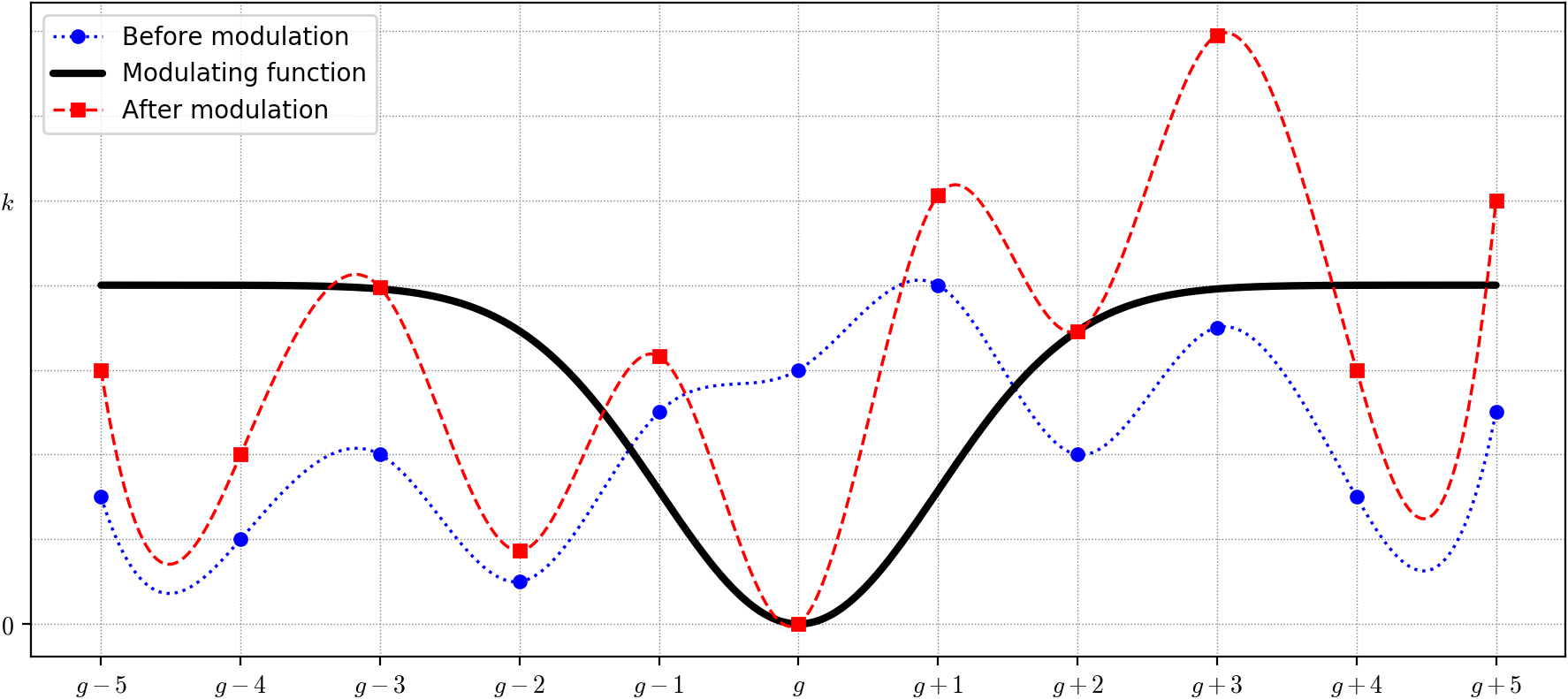}\\
    \end{tabular}
    
    \caption{\textbf{Application of the proposed feature enhancement}. In blue, features $\mathcal{F}$ for pixel $i,j$ in proximity of $d=g_{ij}$, in black the modulating function, in red enhanced features $\mathcal{G}$ for $v_{ij}=1$, applied to correlation features (left) or dissimilarity functions (right).}
    \label{fig:enhancement_dampen}
\end{figure*}

In literature, some techniques were proposed to modify the cost volume of traditional stereo algorithms leveraging prior knowledge such as per-pixel confidence scores \cite{Poggi_2017_ICCV}. A simple, yet effective approach for this purpose consists in hard-replacing matching costs (features, in our case). In \cite{Spyropoulos_2014_CVPR}, matching costs of winning disparities were set to the minimum value and the remaining ones to the maximum, only for those pixels having high confidence score before optimization. The equivalent solution in our domain would consist in zeroing each element corresponding to a disparity $d$ such that $g_{ij} \neq d$. However, this strategy has severe limitations: it is not well-suited for CNNs, either when injected into a pre-trained network -- a large number of zero values would aggressively alter its behavior -- or when plugged during the training from scratch of a new model -- this would cause gradients to not be back-propagated on top of features where the zeros have been inserted. Moreover, no default behavior is defined in case of sub-pixel input disparities, unless they are rounded at the cost of a loss in precision. 

Conversely, we suggest to modulate using a Gaussian function centred on $g_{ij}$, so that the single correlation score or $2F$ features corresponding to the disparity $d = g_{ij}$ are multiplied by the peak of the function, while any other element is progressively multiplied by lower factors, until being dampened the farther they are from $g_{ij}$. Specifically, our modulating function will be

\begin{equation}
k \cdot e^{-\frac{(d-g_{ij})^2}{2c^2}}
\end{equation}
where $c$ determines the width of the Gaussian, while $k$ represents its maximum magnitude and should be greater than or equal to $1$. 
Thus, to obtain a new feature volume $\mathcal{G}$ by multiplying the whole correlation or 3D volume $\mathcal{F}$ regardless of the value of $v_{ij}$, we apply

\begin{equation}
\mathcal{G} = \left(1-v_{ij} + v_{ij} \cdot k \cdot e^{-\frac{(d-g_{ij})^2}{2c^2}}\right) \cdot \mathcal{F}    
\end{equation}
making the weight factor equal to 1 when $v_{ij}=0$.
An example of the effect of our modulation is given in Figure \ref{fig:enhancement_dampen} (left).

\subsection{Applications of guided stereo}
\label{sec:applications}

We will now highlight some notable applications of our technique, that will be exhaustively discussed in the experimental result section.

\textbf{Pre-trained deep stereo networks.} The proposed Gaussian enhancement acts smoothly, yet effectively, on the features already learned by a deep network. Opportunely tuning the hyper-parameters $k$ and $c$, we will prove that our method allows improving the accuracy of pre-trained state-of-the-art networks. 

\textbf{Training from scratch deep stereo networks.} Compared to a brutal zero-product approach, the dampening mechanism introduced by the Gaussian function still allows gradient to flow, making this technique suited for deployment inside a CNN even at training time, so that it can learn from scratch how to better exploit the additional cues. Specifically, the gradients of $\mathcal{G}$ with respect to weights $\mathcal{W}$ will be computed as follows:

\begin{equation}
\begin{split}
\frac{\partial{\mathcal{G}}}{\partial{\mathcal{W}}} 
&= \left(1-v_{ij} + v_{ij} \cdot k \cdot e^{-\frac{(d-g_{ij})^2}{2c^2}}\right) \cdot \frac{\partial{\mathcal{F}}}{\partial{\mathcal{W}}} \\
\end{split}
\end{equation}
Thus, training from scratch a deep network leveraging the sparse input data is possible with our technique.

\textbf{Conventional stereo matching algorithms.} These methods, based on hand-crafted pipelines, can also take advantage of our proposal by leveraging sparse depth cues to improve their accuracy. Sometimes they do not use a similarity measure (\eg, zero mean normalized cross-correlation) to encode the matching cost, for which the same strategy described so far applies, but cost volumes are built using a dissimilarity measure between pixels (\eg, sum of absolute/squared differences or Hamming distance \cite{Secaucus_1994_ECCV}). In both cases, the winning disparity is assigned employing a WTA strategy. When deploying a dissimilarity measure, costs corresponding to disparities close to $g_{ij}$ should be reduced, while the others amplified. We can easily adapt Gaussian enhancement by choosing a modulating function that is the difference between a constant $k$ and a Gaussian function with the same height, obtaining an enhanced volume $\mathcal{G}$ from initial costs $\mathcal{F}$ as



\begin{equation}
\mathcal{G} = \left[1-v_{ij} + v_{ij} \cdot k \cdot \left(1 - e^{-\frac{(d-g_{ij})^2}{2c^2}}\right)\right] \cdot \mathcal{F}
\label{eq:sgm}
\end{equation}
Figure \ref{fig:enhancement_dampen} (right) shows the effect of this formulation.

\section{Experimental Results}
\label{sec:results}

In this section, we report exhaustive experiments proving the effectiveness of the Guided Stereo Matching paradigm showing that the proposed feature enhancement strategy always improves the accuracy of pre-trained or newly trained networks significantly. Moreover, when training the networks from scratch, our proposal increases the ability to generalize to new environments, thus enabling to better tackle domain shifts.
Demo source code is available at \url{https://github.com/mattpoggi/guided-stereo}.

\begin{table}[t]
    \centering
    \scalebox{0.85}
    {
    \begin{tabular}{c|ccc|ccc|}
        \cline{2-7}
         & \multicolumn{3}{c|}{iResNet \cite{Liang_2018_CVPR}} & \multicolumn{3}{c|}{PSMNet \cite{Chang_2018_CVPR}} \\
        \hline
        \multicolumn{1}{|c|}{c} & k=1 & k=10 & k=100 & k=1 & k=10 & k=100\\
        \hline
        \multicolumn{1}{|c|}{0.1} & 2.054 & 1.881 & 2.377 & 4.711 & 4.391 & 4.326 \\
        \multicolumn{1}{|c|}{1} & 1.885 & \textbf{1.338} & 6.857 & 4.540 & \textbf{3.900} & 4.286 \\
        \multicolumn{1}{|c|}{10} & 1.624 & 1.664 & 32.329 & 4.539 & 3.925 & 9.951 \\
        \hline
    \end{tabular}
    }
    \caption{\textbf{Tuning of Gaussian hyper-parameters $k$ and $c$.} Experiments with iResNet (left) and PSMNet (right) trained on synthetic data and tested on KITTI 2015 (average errors without modulation: 1.863 and 4.716)}
    \label{tab:ablation}
\end{table}

\begin{table*}
\centering
\setlength{\tabcolsep}{11pt}
\scalebox{0.85}
{
\begin{tabular}{|c|cc|cc|cccc|c|}
\hline
Model & \multicolumn{2}{|c|}{Training Datasets} & \multicolumn{2}{c|}{Guide} & \multicolumn{4}{c|}{Error rate (\%)} & avg. \\
\cline{2-9} 
 & SceneFlow & KITTI 12 & Train & Test & $>$2 & $>$3 & $>$4 & $>$5 & (px) \\
\hline
\hline
iResNet \cite{Liang_2018_CVPR} & \checkmark & & & & 21.157 & 11.959 & 7.881 & 5.744 & 1.863 \\
iResNet\emph{-gd} & \checkmark & & & \checkmark & 15.146 & 8.208 & 5.348 & 3.881 & 1.431 \\
iResNet\emph{-gd-tr} & \checkmark & & \checkmark & \checkmark & 7.266 & 3.663 & 2.388 & 1.754 & 0.904 \\
\hline
iResNet\emph{-ft} \cite{Liang_2018_CVPR} & \checkmark & \checkmark & & & 9.795 & 4.452 & 2.730 & 1.938 & 1.049 \\
iResNet\emph{-ft-gd} & \checkmark & \checkmark & & \checkmark & 7.695 & 3.812 & 2.524 & 1.891 & 0.994 \\
iResNet\emph{-ft-gd-tr} & \checkmark & \checkmark & \checkmark & \checkmark & \textbf{4.577} & \textbf{2.239} & \textbf{1.476} & \textbf{1.099} & \textbf{0.735} \\
\hline
\hline
PSMNet \cite{Chang_2018_CVPR} & \checkmark & & & & 39.505 & 27.435 & 20.844 & 16.725 & 4.716\\
PSMNet\emph{-gd} & \checkmark & & & \checkmark & 33.386 & 23.125 & 17.598 & 14.101 & 3.900 \\
PSMNet\emph{-gd-tr} & \checkmark & & \checkmark & \checkmark & 12.310 & 3.896 & 2.239 & 1.608 & 1.395 \\
\hline
PSMNet\emph{-ft} \cite{Chang_2018_CVPR} & \checkmark & \checkmark & & & 6.341 & 3.122 & 2.181 & 1.752 & 1.200 \\
PSMNet\emph{-ft-gd} & \checkmark & \checkmark & & \checkmark &  5.707 & 3.098 & 2.266 & 1.842 & 1.092 \\
PSMNet\emph{-ft-gd-tr} & \checkmark & \checkmark & \checkmark & \checkmark & \textbf{2.738} & \textbf{1.829} & \textbf{1.513} & \textbf{1.338} & \textbf{0.763} \\
\hline
\end{tabular}
}
\caption{\textbf{Experimental results on KITTI 2015 dataset \cite{KITTI_2015}.} Tag ``\emph{-gd}'' refers to guiding the network only at test time, ``\emph{-tr}'' to training the model to leverage guide, ``\emph{-ft}'' refers to fine-tuning performed on KITTI 2012 \cite{KITTI_2012}.}
\label{tab:kitti}
\end{table*}

\subsection{Training and validation protocols}

We implemented our framework in PyTorch. For our experiments, we chose two state-of-the-art models representative of the two categories described so far and whose source code is available, respectively iResNet \cite{Liang_2018_CVPR} for correlation-based architectures and PSMNet \cite{Chang_2018_CVPR} for 3D CNNs.
Both networks were pre-trained on synthetic data \cite{Mayer_2016_CVPR} following authors' instructions: 10 epochs for PSMNet \cite{Chang_2018_CVPR} and 650k iterations for iResNet \cite{Liang_2018_CVPR}.  The only difference was the batch size of 3 used for PSMNet since it is the largest fitting in a single Titan Xp GPU used for this research. The proposed guided versions of these networks were trained accordingly following the same protocol.
Fine-tuning on realistic datasets was carried out following the guidelines from the original works when available. In particular, both papers reported results and a detailed training protocol for KITTI datasets \cite{Chang_2018_CVPR,Liang_2018_CVPR}, while training details are not provided for Middlebury \cite{MIDDLEBURY_2014} and ETH3D \cite{ETH3D}, despite results are present on both benchmarks. The following sections will report accurate details about each training protocol deployed in our experiments.
To tune $k$ and $c$, we ran a preliminary experiment applying our modulation on iResNet and PSMNet models trained on synthetic data and tested on KITTI 2015 training set \cite{KITTI_2015}. Table \ref{tab:ablation} shows how the average error varies with different values of $k$ and $c$. According to this outcome, for both networks we will fix $k$ and $c$, respectively, to 10 and 1 for all the following experiments.

To simulate the availability of sparse depth cues, we randomly sample pixels from the ground-truth disparity maps for both training and testing. For this reason, all the evaluation will be carried out on the training splits available from KITTI, Middlebury and ETH3D datasets. Finally, we point out that the KITTI benchmarks include a depth completion evaluation. However, it aims at assessing the performance of monocular camera systems coupled with an active sensor (\ie, LiDAR) and thus the benchmark does not provide the stereo pairs required for our purposes. 

\subsection{Evaluation on KITTI}
\label{sec:kitti}

At first, we assess the performance of our proposal on the KITTI 2015 dataset \cite{KITTI_2015}. Table \ref{tab:kitti} collects results obtained with iResNet and PSMNet trained and tested in different configurations. For each experiment we highlight the imagery used during training, respectively the SceneFlow dataset alone \cite{Mayer_2016_CVPR} or the KITTI 2012 \cite{KITTI_2012} used for fine-tuning (``\emph{-ft}''). Moreover, we report results applying our feature enhancement to pre-trained networks (\ie, at test time only, ``\emph{-gd}'') and training the networks from scratch  (``\emph{-gd-tr}''). 
For each experiment, we report the error rate as the percentage of pixels having a disparity error larger than a threshold, varying between 2 and 5, as well as the mean average error on all pixels with available ground-truth.
To obtain sparse measurements, we randomly sample pixels with a density, computed on the whole image, of 5\% on SceneFlow \cite{Mayer_2016_CVPR}. On KITTI, we keep a density of 15\%, then remove unlabelled pixels to obtain again 5\% with respect to the lower portion of the images with available ground-truth (i.e., a $220\times 1240$ pixel grid). 

\begin{figure*}
    \centering
    \renewcommand{\tabcolsep}{1pt}   
    \begin{tabular}{ccc}

            \includegraphics[width=0.32\textwidth]{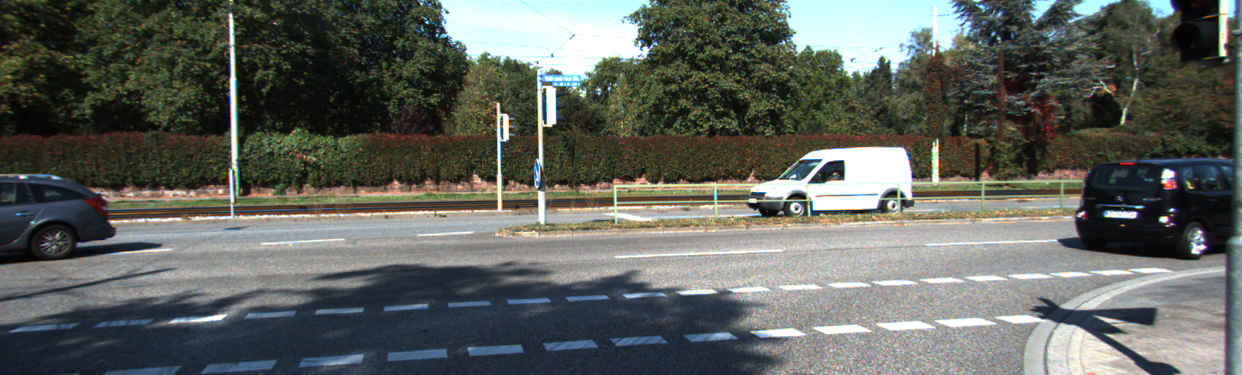} &
            \begin{overpic}[width=0.32\textwidth]{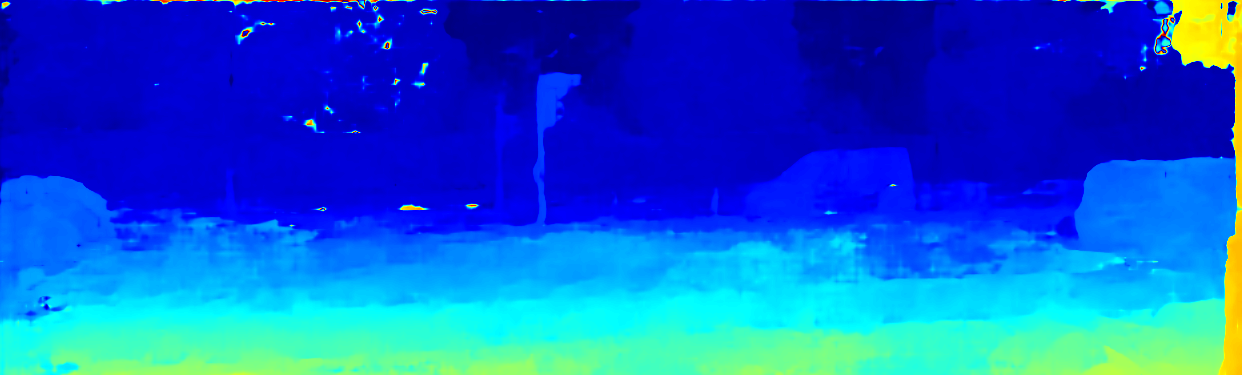}
            \put (2,25) {$\displaystyle\textcolor{white}{\textbf{avg = 2.82}}$}
            \put (2,18) {$\displaystyle\textcolor{white}{\textbf{$>3$ = 27.6\%}}$}            
            \end{overpic} &
            \begin{overpic}[width=0.32\textwidth]{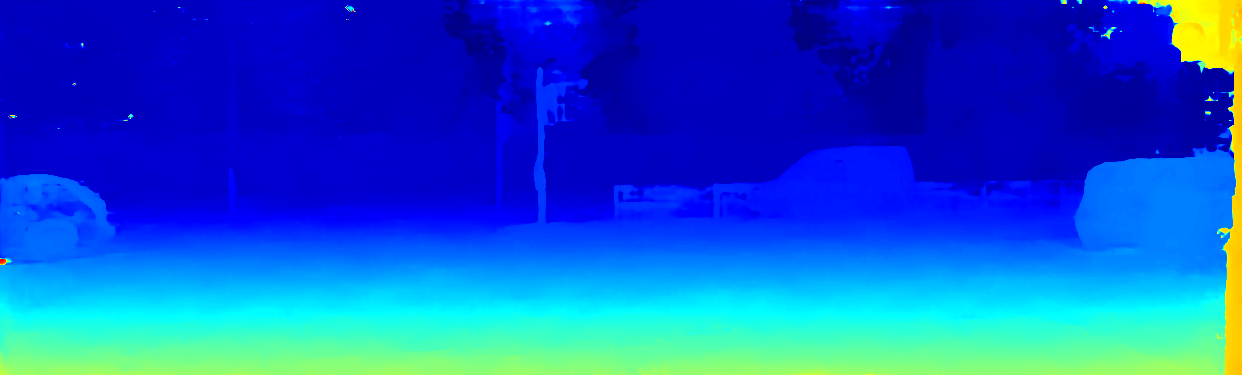}
            \put (2,25) {$\displaystyle\textcolor{white}{\textbf{avg = 1.56}}$}
            \put (2,18) {$\displaystyle\textcolor{white}{\textbf{$>3$ = 3.0\%}}$}                
            \end{overpic} \\
    \end{tabular}
    \caption{\textbf{Comparison between variants of PSMNet \cite{Chang_2018_CVPR}.} From top to bottom, reference image from 000022 pair (KITTI 2015 \cite{KITTI_2015}), disparity maps obtained by PSMNet \cite{Chang_2018_CVPR} and PSMNet\emph{-gd-tr}, both trained on synthetic images only.}
    \label{fig:000022}
\end{figure*}

From Table \ref{tab:kitti} we can notice how both baseline architectures (row 1 and 7) yield large errors when trained on SceneFlow dataset only. In particular, PSMNet seems to suffer the domain shift more than correlation-based technique iResNet. By applying the proposed feature enhancement to both networks, we can ameliorate all metrics sensibly, obtaining first improvements to network generalization capability. In particular, by looking at the $>3$ error rate, usually taken as the reference metric in KITTI, we have an absolute reduction of about 3.8 and 4.3 \% respectively for iResNet\emph{-gd} and PSMNet\emph{-gd} compared to the baseline networks.
In this case, we point out once more that we are modifying only the features of a pre-trained network, by just altering what the following layers are used to process. Nonetheless, our proposal preserves the learned behavior of the baseline architecture increasing at the same time its overall accuracy.

When training the networks from scratch to process sparse inputs with our technique, iResNet\emph{-gd-tr} and PSMNet\emph{-gd-tr} achieve a notable drop regarding error rate and average error compared to the baseline models. Both reach degrees of accuracy comparable to those of the original network fine-tuned on KITTI 2012, iResNet\emph{-ft} and PSMNet\emph{-ft} without actually being trained on such realistic imagery, by simply exploiting a small amount of depth samples (about 5\%) through our technique.
Moreover, we can also apply the feature enhancement paradigm to fine-tuned models. From Table \ref{tab:kitti} we can notice again how our technique applied to the fine-tuned models still improves their accuracy. Nonetheless, fine-tuning the networks pre-trained to exploit feature enhancement leads to the best results across all configurations, with an absolute decrease of about 2.2 and 1.3\% compared, respectively, to the already low error rate of iResNet\emph{-ft} and PSMNet\emph{-ft}. 
Finally, Figure \ref{fig:000022} shows a comparison between the outputs of different PSMNet variants, highlighting the superior generalization capacity of PSMNet\emph{-gd-tr} compared to baseline model.

\begin{table*}
\centering
\setlength{\tabcolsep}{11pt}
\scalebox{0.85}
{
\begin{tabular}{|c|cc|cc|cccc|c|}
\hline
Model & \multicolumn{2}{|c|}{Training Datasets} & \multicolumn{2}{c|}{Guide} & \multicolumn{4}{c|}{Error rate (\%)} & avg. \\
\cline{2-9} 
 & SceneFlow & trainingQ & Train & Test & $>$0.5 & $>$1 & $>$2 & $>$4 & (px) \\
\hline
\hline
iResNet \cite{Liang_2018_CVPR} & \checkmark & & & & 69.967 & 50.893 & 30.742 & 16.019 & 2.816  \\
iResNet\emph{-gd} & \checkmark & & & \checkmark & 62.581 & 40.831 & 22.154 & 10.889 & 2.211 \\
iResNet\emph{-gd-tr} & \checkmark & & \checkmark & \checkmark & 44.385 & 25.555 & 12.505 & 5.776 & 1.470  \\
\hline
iResNet\emph{-ft} \cite{Liang_2018_CVPR} & \checkmark & \checkmark & & & 69.526 & 49.027 & 28.178 & 14.126 & 2.682  \\
iResNet\emph{-ft-gd} & \checkmark & \checkmark & & \checkmark & 60.979 & 36.255 & 19.558 & 10.136 & 2.130 \\
iResNet\emph{-ft-gd-tr} & \checkmark & \checkmark & \checkmark & \checkmark & \textbf{31.526} & \textbf{17.045} & \textbf{8.316} & \textbf{4.307} & \textbf{0.930} \\
\hline
\hline
PSMNet \cite{Chang_2018_CVPR} & \checkmark & & & & 54.717 & 33.603 & 20.239 & 13.304 & 5.332 \\
PSMNet\emph{-gd} & \checkmark & & & \checkmark & 53.090 & 31.416 & 18.619 & 12.588 & 4.921 \\
PSMNet\emph{-gd-tr} & \checkmark & & \checkmark & \checkmark & 83.433 & 54.147 & 7.472 & 3.208 & 1.732 \\
\hline
PSMNet\emph{-ft} \cite{Chang_2018_CVPR} & \checkmark & \checkmark & & & 45.523 & 25.993 & 15.203 & 8.884 & 1.964 \\
PSMNet\emph{-ft-gd} & \checkmark & \checkmark & & \checkmark & 44.004 & 25.151 & 14.337 & 8.676 & 1.894 \\
PSMNet\emph{-ft-gd-tr} & \checkmark & \checkmark & \checkmark & \checkmark & \textbf{32.715} & \textbf{15.724} & \textbf{6.937} & \textbf{3.756} & \textbf{1.348} \\
\hline
\end{tabular}
}
\caption{\textbf{Experimental results on Middlebury v3 \cite{MIDDLEBURY_2014}.} ``\emph{-gd}'' refers to guiding the network only at test time, ``\emph{-tr}'' to training the model to leverage guide, ``\emph{-ft}'' refers to fine-tuning performed on \emph{trainingQ} split.}
\label{tab:middlebury}
\end{table*}

\begin{table*}
\centering
\setlength{\tabcolsep}{11pt}
\scalebox{0.85}
{
\begin{tabular}{|c|cc|cc|cccc|c|}
\hline
Model & \multicolumn{2}{|c|}{Training Datasets} & \multicolumn{2}{c|}{Guide} & \multicolumn{4}{c|}{Error rate (\%)} & avg. \\
\cline{2-9} 
 & SceneFlow & ETH3D & Train & Test & $>$0.5 & $>$1 & $>$2 & $>$4 & (px) \\
\hline
\hline

iResNet \cite{Liang_2018_CVPR} & \checkmark & & & & 57.011 & 36.944 & 20.380 & 12.453 & 5.120 \\
iResNet\emph{-gd} & \checkmark & & & \checkmark & 50.361 & 29.767 & 16.495 & 10.293 & 2.717 \\
iResNet\emph{-gd-tr} & \checkmark & & \checkmark & \checkmark & 26.815 & 10.673 & 3.555 & 1.312 & 0.537 \\
\hline 
iResNet\emph{-ft} \cite{Liang_2018_CVPR} & \checkmark & \checkmark & & & 48.360 & 26.212 & 11.865 & 4.678 & 0.997  \\
iResNet\emph{-ft-gd} & \checkmark & \checkmark & & \checkmark & 47.539 & 22.639 & 8.153 & 2.445 & 0.820 \\
iResNet\emph{-ft-gd-tr} & \checkmark & \checkmark & \checkmark & \checkmark & \textbf{23.433} & \textbf{8.694} & \textbf{2.803} & \textbf{0.876} & \textbf{0.443} \\
\hline
\hline
PSMNet \cite{Chang_2018_CVPR} & \checkmark & & & & 45.522 & 23.936 & 12.550 & 7.811 & 5.078 \\
PSMNet\emph{-gd} & \checkmark & & & \checkmark & 43.667 & 21.140 & 10.773 & 7.081 & 4.739 \\
PSMNet\emph{-gd-tr} & \checkmark & & \checkmark & \checkmark & 96.976 & 71.970 & 2.730 & 0.512 & 1.266 \\
\hline
PSMNet\emph{-ft} \cite{Chang_2018_CVPR} & \checkmark & \checkmark & & & 28.560 & 11.895 & 4.272 & 1.560 & 0.560 \\
PSMNet\emph{-ft-gd} & \checkmark & \checkmark & & \checkmark & 25.707 & 10.095 & 3.084 & 1.123 & 0.505 \\
PSMNet\emph{-ft-gd-tr} & \checkmark & \checkmark & \checkmark & \checkmark & \textbf{17.865} & \textbf{4.195} & \textbf{1.360} & \textbf{0.817} & \textbf{0.406} \\
\hline
\end{tabular}
}
\caption{\textbf{Experimental results on ETH3D dataset \cite{ETH3D}.} ``\emph{-gd}'' refers to guiding the network only at test time, ``\emph{-tr}'' to training the model to leverage guide, ``\emph{-ft}'' refers to fine-tuning performed on the training split (see text).}
\label{tab:eth}
\end{table*}

\subsection{Evaluation on Middlebury}
\label{sec:middlebury}
    
We also evaluated our proposal on Middlebury v3 \cite{MIDDLEBURY_2014}, since this dataset is notoriously more challenging for end-to-end architectures because of the very few images available for fine-tuning and the more heterogeneous scenes framed compared to KITTI. Table \ref{tab:middlebury} collects the outcome of these experiments. We use the same notation adopted for KITTI experiments. All numbers are obtained processing the \emph{additional} split of images at quarter resolution, since higher-resolution stereo pairs do not fit into the memory of a single Titan Xp GPU. Fine-tuning is carried out on the \emph{training} split. We compute error rates with thresholds 0.5, 1, 2 and 4 as usually reported on the online benchmark. Sparse inputs are randomly sampled with a density of 5\% from ground-truth data.
We can notice how applying feature enhancement on both pre-trained models or training new instances from scratch gradually reduces the errors as observed on KITTI experiments. Interestingly, we point out that while this trend is consistent for iResNet\emph{-gd} and iResNet\emph{-gd-tr}, a different behavior occurs for PSMNet\emph{-gd-tr}. In particular, we can notice a huge reduction of the error rate setting the threshold to $>2$ and $>4$. On the other hand, the percentage of pixels with lower disparity errors $>0.5$ and $>1$ gets much higher. Thus, with PSMNet, the architecture trained with guiding inputs seems to correct most erroneous patterns at the cost of increasing the number of small errors. Nonetheless, the average error always decreases significantly. 

Regarding fine-tuning, we ran about 300 epochs for each baseline architecture to obtain the best results. After this phase, we can observe minor improvements for iResNet, while PSMNet improves its accuracy by a large margin. Minor, but consistent improvements are yielded by iResNet\emph{-ft-gd} and PSMNet\emph{-ft-gd}. Finally, we fine-tuned iResNet\emph{-ft-gd-tr} and PSMNet\emph{-ft-gd-tr} for about 30 epochs, sufficient to reach the best performance. Again, compared to all other configurations, major improvements are always yielded by guiding both networks. 

\subsection{Evaluation on ETH3D}

Finally, we assess the performance of our method on the ETH3D dataset \cite{ETH3D}. In this case we split the training dataset, using images from \emph{delivery\_area\_1l}, \emph{delivery\_area\_1s}, \emph{electro\_1l}, \emph{electro\_1s}, \emph{facade\_1s}, \emph{forest\_1s}, \emph{playground\_1l}, \emph{playground\_1s}, \emph{terrace\_1s}, \emph{terrains\_1l}, \emph{terrains\_1s} for fine-tuning and the remaining ones for testing. For \emph{-ft} models, we perform the same number of training epochs as for Middlebury dataset, and Table \ref{tab:eth} collects results for the same configurations considered before. We can notice behaviors similar to what reported in previous experiments. By guiding iResNet we achieve major improvements, nearly halving the average error, while the gain is less evident for PSMNet, although beneficial on all metrics. Training iResNet\emph{-gd-tr} and PSMNet\emph{-gd-tr} leads to the same outcome noticed during our experiments on Middlebury. In particular, PSMNet\emph{-gd-tr} still decimates the average error and percentage of errors greater than 2 and 4 at the cost of a large amount of pixels with error greater than 0.5 and 1.
With this dataset, fine-tuning the baseline models enables to significantly increase the accuracy of both, in particular decimating the average error from beyond 5 pixels to less than 1. Nevertheless, our technique is useful even in this case, enabling minor yet consistent improvements when used at test time only and significant boosts for iResNet\emph{-ft-gd-tr} and PSMNet\emph{-ft-gd-tr}, improving all metrics by a large margin.

\subsection{Evaluation with SGM}

To prove the effectiveness of our proposal even with conventional stereo matching algorithms, we evaluated it with SGM \cite{hirschmuller2005accurate}. For this purpose, we used the code provided by \cite{rsgm}, testing it on all datasets considered so far. As pointed out in Section \ref{sec:applications}, feature enhancement can be opportunely revised as described in Equation \ref{eq:sgm} to deal with dissimilarity measures. We apply this formulation before starting scanline optimizations.
As for any previous experiments, we sample sparse inputs as described in Section \ref{sec:results}, obtaining an average density below 5\%.  Table \ref{tab:sgm-kitti} reports the comparison between SGM and its \emph{cost enhanced} counterpart using sparse input cues (SGM-\emph{gd}) on KITTI 2012 \cite{KITTI_2012} (top) and KITTI 2015 \cite{KITTI_2015} (bottom). With both datasets we can notice dramatic improvements in all metrics. In particular, the amount of outliers $>2$ is more than halved and reduced in absolute by about 4, 3 and 2\% for higher error bounds.
Table \ref{tab:sgm-more} reports experiments on Middlebury \emph{training} (top) and \emph{additional} (bottom) splits \cite{MIDDLEBURY_2014}, as well as on the entire ETH3D training set \cite{ETH3D}. Experiments on Middlebury are carried out at quarter resolution for uniformity with previous experiments with deep networks reported in Section \ref{sec:middlebury}. The error rate is reduced by about 5.5\% for $>0.5$ on the three experiments, by about 7.5\% for $>1$, nearly halved on Middlebury and reduced by a factor 2.5 on ETH3D for $>2$, and  by 6, 6 and 3\% for $>4$. Finally, average errors are reduced by 1.1 on Middlebury and 1.4 on ETH3D.

\begin{table}
\centering
\setlength{\tabcolsep}{8.5pt}
\scalebox{0.85}
{
\begin{tabular}{|c|cccc|c|}
\hline
Alg. & \multicolumn{4}{c|}{Error rate (\%)} & avg. \\
\cline{2-5} 
 & $>$2 & $>$3 & $>$4 & $>$5 & (px) \\
\hline
\hline
SGM \cite{hirschmuller2005accurate} & 11.845 & 8.553 & 7.109 & 6.261 & 2.740\\
SGM\emph{-gd} & \textbf{5.657} & \textbf{4.601} & \textbf{4.162} & \textbf{3.892} & \textbf{2.153}\\
\hline
\hline
SGM \cite{hirschmuller2005accurate} & 15.049 & 8.843 & 6.725 & 5.645 & 2.226\\
SGM\emph{-gd} & \textbf{6.753} & \textbf{4.294} & \textbf{3.625} & \textbf{3.282} & \textbf{1.680}\\
\hline
\end{tabular}
}
\caption{\textbf{Experimental results on KITTI.} Comparison between raw \cite{hirschmuller2005accurate} and guided SGM on KITTI 2012 (top) and 2015 (bottom).}
\label{tab:sgm-kitti}
\end{table}

\begin{table}
\centering
\setlength{\tabcolsep}{7.5pt}
\scalebox{0.85}
{
\begin{tabular}{|c|cccc|c|}
\hline
Alg. & \multicolumn{4}{c|}{Error rate (\%)} & avg. \\
\cline{2-5} 
 & $>$0.5 & $>$1 & $>$2 & $>$4 & (px) \\
\hline
\hline
SGM \cite{hirschmuller2005accurate} & 62.428 & 32.849 & 20.620 & 15.786 & 4.018\\
SGM\emph{-gd} & \textbf{56.882} & \textbf{24.608} & \textbf{12.655} & \textbf{9.909} & \textbf{2.975}\\
\hline
\hline
SGM \cite{hirschmuller2005accurate} & 64.264 & 31.966 & 18.741 & 14.675 & 4.978\\
SGM\emph{-gd} & \textbf{59.596} & \textbf{24.856} & \textbf{11.307} & \textbf{8.960} & \textbf{3.815}\\
\hline
\hline
SGM \cite{hirschmuller2005accurate} & 58.994 & 27.356 & 10.685 & 5.632 & 1.433\\
SGM\emph{-gd} & \textbf{54.051} & \textbf{20.156} & \textbf{4.169} & \textbf{2.459} & \textbf{1.032}\\
\hline
\end{tabular}
}
\caption{\textbf{Experimental results on Middlebury v3 and ETH3D datasets.} Comparison between raw \cite{hirschmuller2005accurate} and guided SGM on \emph{training} (top), \emph{additional} (middle) \cite{MIDDLEBURY_2014} and ETH3D \cite{ETH3D} (bottom).}
\label{tab:sgm-more}
\end{table}

The evaluation with SGM highlights how our technique can be regarded as a general purpose strategy enabling notable improvements in different contexts, ranging from state-of-the-art deep learning frameworks to traditional stereo algorithms.

\begin{table}
\centering
\setlength{\tabcolsep}{11pt}
\scalebox{0.7}
{
\begin{tabular}{|c|cc|cc|}
\hline
Model / Alg. & \multicolumn{2}{c|}{$<$2\%} & \multicolumn{2}{c|}{avg.} \\
\cline{2-5}
 & All & NoG & All & NoG \\
\hline
iResNet \cite{Liang_2018_CVPR} & 18.42 & 18.37 & 1.28 & 1.28 \\
iResNet+Martins \etal \cite{martins2018fusion} & 18.14 & 18.09 & 1.26 & 1.26 \\
iResNet+Marin \etal (opt.) & 15.20 & 18.37 & 1.07 & 1.28 \\
iResNet\emph{-gd} & 11.12 & 10.99 & 1.04 & 1.03 \\
iResNet\emph{-gd-tr} & \textbf{5.38} & \textbf{5.27} & \textbf{0.77} & \textbf{0.77} \\
\hline
iResNet\emph{-ft} \cite{Liang_2018_CVPR} & 5.29 & 5.30 & 0.81 & 0.81 \\
iResNet\emph{-ft}+Martins \etal \cite{martins2018fusion} & 5.26 & 5.28 & 0.80 & 0.80 \\
iResNet\emph{-ft}+Marin \etal \cite{Marin_ECCV_2016} (opt.) & 4.48 & 5.30 & 0.67 & 0.81 \\
iResNet\emph{-ft-gd} & 3.14 & 3.13 & 0.64 & 0.64 \\
iResNet\emph{-ft-gd-tr} & \textbf{1.91} & \textbf{1.88} & \textbf{0.55} & \textbf{0.55} \\
\hline
PSMNet \cite{Chang_2018_CVPR} & 38.60 & 38.86 & 2.36 &  2.37 \\
PSMNet+Martins \etal \cite{martins2018fusion} & 38.32 & 38.58 & 2.33 & 2.34 \\
PSMNet+Marin \etal \cite{Marin_ECCV_2016} (opt.) & 34.85 & 38.86 & 1.99 &  2.17 \\
PSMNet\emph{-gd} & 33.47 & 33.74 & 2.07 & 2.08 \\
PSMNet\emph{-gd-tr} & \textbf{21.57} & \textbf{21.30} & \textbf{1.60} & \textbf{1.59} \\
\hline
PSMNet-ft \cite{Chang_2018_CVPR} & 1.71 & 1.73 & 0.72 & 0.72\\
PSMNet\emph{-ft}+Martins \etal \cite{martins2018fusion} & 1.82 & 1.83 & 0.72 & 0.72 \\
PSMNet\emph{-ft}+Marin \etal \cite{Marin_ECCV_2016} (opt.) & 1.52 & 1.73 & 0.66 &  0.72 \\
PSMNet\emph{-ft-gd} & 1.13 & 1.15 & 0.60 & 0.61 \\
PSMNet\emph{-ft-gd-tr} & \textbf{0.67} & \textbf{0.67} & \textbf{0.47} & \textbf{0.47} \\
\hline
SGM \cite{hirschmuller2005accurate} & 9.42 & 9.54 & 1.24 & 1.24 \\
SGM+Martins \etal \cite{martins2018fusion} & 9.41 & 9.53 & 1.24 & 1.24 \\
SGM+Marin \etal \cite{Marin_ECCV_2016} (opt.) & 8.15 & 9.54 & 1.14 & 1.24 \\
SGM\emph{-gd} & \textbf{2.79} & \textbf{3.03} & \textbf{0.99} & \textbf{0.99} \\
\hline
\end{tabular}
}
\caption{Experiments on KITTI Velodyne, seq. \emph{2011\_09\_26\_0011}.}
\label{tab:rebuttal}
\end{table}

\subsection{Experiments with Lidar measurements}

Finally, we evaluate the proposed paradigm using as guide the raw and noisy measurements from a Velodyne sensor \cite{VELODYNE}, to underline the practical deployability of the proposed solution further. Table \ref{tab:rebuttal} reports experiment from sequence 2011\_09\_26\_0011 of the KITTI raw dataset \cite{KITTI_RAW}. We compare our framework with fusion strategies proposed by Martins \etal \cite{martins2018fusion} and Marin \etal \cite{Marin_ECCV_2016}, combining outputs by the stereo networks respectively with monocular estimates (using the network by Guo \etal \cite{Guo_2018_ECCV}) and Lidar, reporting the ideal result as in \cite{Marin_ECCV_2016}. Ground-truth labels for evaluation are provided by \cite{Uhrig2017THREEDV}. Our proposal consistently outperforms fusion approaches by a large margin, evaluating on all pixels (All) as well as excluding those with Lidar (NoG) to stress that the improvement yielded by our method is not limited to pixels with associated Lidar measurement in contrast to fusion techniques \cite{Marin_ECCV_2016}.

\section{Conclusions}

In this paper, we proposed Guided Stereo Matching, a novel paradigm to boost state-of-the-art deep architectures trained for dense disparity inference using as additional input cue a small set of sparse depth measurements provided by an external source. By enhancing the features that encode matching relationships between pixels across left and right images, we can improve the accuracy and robustness to domain shifts. Our feature enhancement strategy can be used seamlessly with pre-trained models, yielding significant accuracy improvements. More importantly, thanks to its fully-differentiable nature, it can even be used to train new instances of a CNN from scratch, in order to fully exploit the input guide and thus to remarkably improve overall accuracy and robustness to domain shifts of deep networks. 
Finally, our proposal can be deployed even with conventional stereo matching algorithms such as SGM, yielding significant improvements as well. 
The focus of future work will be on devising strategies to guide our method without relying on active sensors. For instance, selecting reliable depth labels leveraging confidence measures \cite{Poggi_2017_ICCV} -- since this strategy proved to be successful for self-supervised adaptation \cite{Tonioni_2017_ICCV,Tonioni_2019_CVPR} and training learning-based confidence measures \cite{Tosi_2017_BMVC} -- or from the output of a visual stereo odometry systems \cite{Wang_2017_ICCV}. 

\textbf{Acknowledgement.} We gratefully acknowledge the support of NVIDIA Corporation with the donation of the Titan Xp GPU used for this research.

{\small
\bibliographystyle{ieee}
\bibliography{stereo}
}

\end{document}